# A Semantic Web Framework for Automated Smart Assistants: COVID-19 Case Study

YUSUF SERMET*

IIHR – Hydroscience & Engineering, University of Iowa, Iowa City, IA, USA

IBRAHIM DEMIR

Civil and Environmental Engineering, University of Iowa, Iowa City, IA, USA

COVID-19 pandemic elucidated that knowledge systems will be instrumental in cases where accurate information needs to be communicated to a substantial group of people with different backgrounds and technological resources. However, several challenges and obstacles hold back the wide adoption of virtual assistants by public health departments and organizations. This paper presents the *Instant Expert*, an open-source semantic web framework to build and integrate voice-enabled smart assistants (i.e. chatbots) for any web platform regardless of the underlying domain and technology. The component allows non-technical domain experts to effortlessly incorporate an operational assistant with voice recognition capability into their websites. Instant Expert is capable of automatically parsing, processing, and modeling Frequently Asked Questions pages as an information resource as well as communicating with an external knowledge engine for ontology-powered inference and dynamic data utilization. The presented framework utilizes advanced web technologies to ensure reusability and reliability, and an inference engine for natural language understanding powered by deep learning and heuristic algorithms. A use case for creating an informatory assistant for COVID-19 based on the Centers for Disease Control and Prevention (CDC) data is presented to demonstrate the framework's usage and benefits.

## 1 INTRODUCTION

The immense amount of data is constantly being generated in a variety of fields including environmental, biological, and physical sciences due to the rapid advancements in monitoring and computational techniques [1][2]. The massive data requires efficient tools and intermediates for data management [3], analysis, visualization, and communication [4]. Web-based information systems (IS) serve as one-stop platforms to access, analyze, and explore information effectively for decision-making purposes [5][6]. Although current systems are proved to be successful to communicate and analyze data efficiently, they still suffer from the complexity and the higher learning curves due to the limitations of conventional interaction methods [7]. Users of information systems (e.g. public, workers, managers, decision-makers, organizational leaders) often look for a certain piece of knowledge for which they may have to master the functionalities and resources provided by the IS [8]. This is especially tedious and discouraging for users who are not continuous visitors to the system. Thus, modern approaches are needed to free the users from the nuances and complexities of information systems and provide a feasible tool to access knowledge [9]. Automated knowledge communication will be critical for participatory decision-making process in disaster mitigation as well [10][11].

---

* Corresponding author: msermet@uiowa.edu

With the advancements in artificial intelligence, there is a significant surge in research of chatbots, which can be defined as intelligent agents (i.e. assistants) that have the ability to comprehend natural language queries and produce a direct and factual response utilizing data and service providers [12]. More recently, the increased prevalence of deep learning tools and generalized algorithms resulted in the growth of chatbots fueled from deep learning's ability to make robust and scalable inferences out of high-dimensional and likely noisy data [13]. Technology companies have been taking the lead on operational virtual assistants integrated into their ecosystem which triggered a brand new and massive market that was valued at US$ 2.2 billion in 2018 and is expected to reach US$ 11.3 billion by 2024 [14]. However, the usage of chatbots for effective and reliable information communication is not widespread among public, government, and scientific communities [15]. As of 2018, only 16% of Internet users have ever used a chatbot [16] despite the fact that there is a substantial demand [17]. Several publications emphasize the potential chatbots hold to serve as the next generation information communication tool and make the case for an urgent need for chatbot development and adoption in their respective fields [18][19]. The United States Army Corps of Engineers (USACE) Institute for Water Resources reports that virtual assistants hold great promise for the USACE to better serve the public, enhance workflow, and improve decision-making while saving time and financial resources [20]. Voice-enabled knowledge systems can support novel virtual reality systems [21] for emergency response training and disaster risk communication [22]. More recently, the rise of the COVID-19 pandemic made it clear that chatbots will be instrumental in cases where accurate information needs to be communicated to a substantial group of people with different backgrounds and technological resources as reported in a Nature article [23]. Sohrabi et al. [24] describe how artificial intelligence-powered chatbots can be widely used to provide up-to-date information regarding the COVID-19 out-break in the context of prevention and management.

While these reports put forth a futuristic and essential vision, they are held back by several challenges regarding the adoption of virtual assistants [25]. One of the major obstacles of virtual assistant development and utilization is that the available services usually require a centralized data flow (e.g. server-side processing of voice, sentence, or data) involving a third-party service provider which presents concerns in terms of data privacy, security, and compatibility [26]. Currently available frameworks for chatbot development often employ vertical integration in which the developers find themselves stuck with the ecosystem of the service provider [27]. A 2019 report on the virtual personal assistant market concludes that the high-growth inflection point of the market and wide adoption of smart assistants in many fields will occur as open-source and free assistant development tools [27]. A second major challenge is that the workforce and / or financial resources required to initialize and maintain a chatbot discourages establishments, governments, research groups, and non-profits, while for larger companies and organizations these costs justify themselves for the long term [28]. Furthermore, active internet connection is often needed for chatbot to operate at its simplest level, which presents a significant setback as many use cases do not have a reliable and consistent internet connection [20].

This paper presents Instant Expert, an open-source semantic web component to build and integrate voice-enabled smart assistants (i.e. chatbots) for any web platform regardless of the underlying domain and technology. The component allows non-technical domain experts to incorporate an operational assistant with voice recognition capability into their websites by simply adding as little as a single line of HTML code while customizations are enabled for more advanced use cases. The component entails an encapsulated user interface, that accepts natural language questions via text and speech inputs as well as selection from a predefined list of questions. A knowledge generation module processes questions to map them to the configured data resource and returns the answers utilizing its inference engine and natural language mapping methods powered by deep learning and heuristic algorithms. Instant Expert is capable of automatically parsing, processing, and modeling internal (same-origin) or external (cross-origin) Frequently Asked Questions (FAQ) webpages as an information resource as well as communicating with an external knowledge engine for more advanced use cases. The presented framework is powered by advanced web technologies to ensure reusability and reliability and deep learning to perform accurate natural language mappings. A use case for creating an informatory assistant for COVID-19 based on the Centers for Disease Control and Prevention (CDC) data is presented to demonstrate the framework's usage and benefits.

The main contributions of this research can be summarized as follows. The presented component makes it possible for any web system on any domain to have its own voice-enabled smart assistant to instantly provide factual responses to natural language queries. It can grow the system's visibility and increase user retention and satisfaction due to providing the user with the information they desire without a hassle. The component can especially be valuable for individual developers, academic research groups, small companies, non-profits, and public offices that may not have the resources for the development and maintenance of commercial smart assistants for their organization. The framework liberates developers from the limitations and boundaries of any given ecosystem and maximizes customization based on available resources and needs while providing a generalized framework to offer standardized, robust, and efficient smart assistants. The framework is completely built on web technologies (e.g. HTML5, JavaScript, CSS) working on the client-side which eliminates the dependence to server-side technologies and assures data privacy. Finally, the Instant Expert's independence from any service pro-vide and ecosystem paves the way for its expansion into virtual reality (e.g. A-Frame), and augmented reality (i.e. HoloLens and Magic Leap) applications through web [29].

The remainder of this article is organized as follows. Section 2 presents the methodology of the development and implementation of the intelligent web framework. Section 3 describes a case study on COVID-19 and pro-vides benchmark results and performance analysis. Section 4 concludes the articles with a summary of contributions and future work.

## 2 METHODS

The Instant Expert web component entails several components all of which are implemented using Hypertext Markup Language 5 (HTML5), JavaScript (JS), and Cascading Style Sheets (CSS). The component runs completely on the client-side (i.e. utilizing only the client's hardware) by which significantly minimizes the workload, and consequently, the maintenance cost to employ a smart assistant. The framework can be abstracted into two main semantic units: 1) user interaction and interface and 2) knowledge generation. Various features of ECMAScript 6 (ES6) have been utilized in the software including JS Modules to avoid global namespace pollution and provide a degree of encapsulation against potential conflicts between the smart assistant and hosting web platform. The framework can be imported into any web application as a script as demonstrated below.

```
<script src="instant-expert-dist.js"></script>
```

Additional benefits of utilizing ES6 include default parameter values, string interpolation, OOP-style (Object-Oriented Programming) classes, which lead to a robust, clean, and extendible product. The usage of web frameworks (e.g. Polymer, Stencil) is considered when designing the web component, though consciously avoided to minimize the learning curve and eliminate the dependency on any framework's abstraction. The code of Instant Expert is fully open-source and accessible via GitHub (https://github.com/uihilab/instant-expert) with detailed documentation guiding developers on how to employ and configure the framework. Figure 1 summarizes the overall architecture of the presented framework.

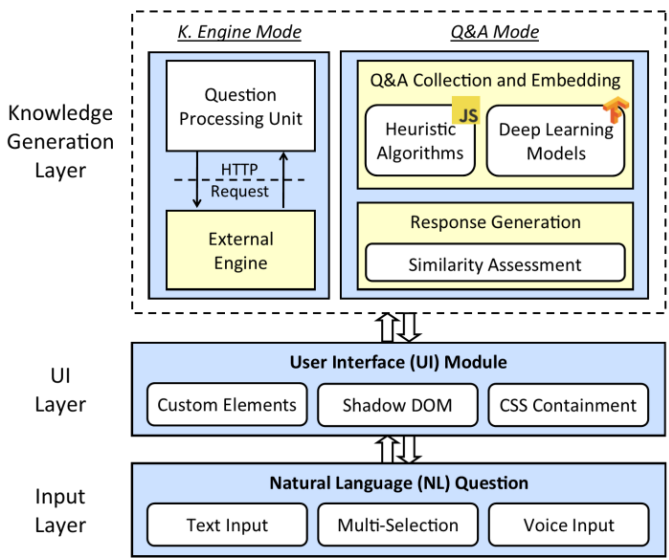

Figure 1: System architecture of the Instant Expert

## 2.1 User Interaction and Interface

The Instant Expert is built upon the features provided by the Web Components. Web Components are a collection of web technologies combined with the purpose of creating reusable, customizable, and encapsulated HTML elements [30]. It is mainly powered by three web technologies. Custom Elements enables the creation of new HTML elements. HTML templates offer the mechanism to define HTML content that can be instantiated during runtime instead of getting rendered when the page is loaded. Shadow DOM provides encapsulation of an element's features with a shadow tree associated with the web component [31]. The major consideration when designing the component is to prevent it to affect the visual and functional integrity of the hosting web site, as well as to prevent it from getting affected by the hosting web site. Shadow DOM prevents potential integration complications by creating a shadow root under the custom HTML element and rendering it separately from the main document DOM. These new technologies are not yet widely adopted in the industry and especially in academia despite the remarkable benefits they offer. Another aim of this paper is to serve as an example to advantages of Web Components by reducing the technicality.

### CSS Containment

Containment is a CSS feature that aims to isolate a contained element's contents from the rest of the document, as much as possible. The main purpose of the containment is to provide optimization and offer stability in rendering and painting of web pages by providing a standardized way. Browsers are always looking to make optimizations when rendering a page. Containment provides a standardized way to tell the browser where and how it can optimize without breaking the intended functionality. When using third-party DOM, such as the Instant Expert, containment can prove to be useful to sandbox the component to protect and increase the performance of the page. It should be noted that containment is not a security feature and is not aimed to provide a full encapsulation.

There are 4 main types of containment that can either be used individually or in groups. The details of each containment type can be found in the specification document published by the World Wide Web Consortium (W3C) [32]. For the purposes of Instant Expert, Content Containment has been used which combines Layout, Paint, and Style Containments. A web component already brings the containing functionality; however, Content Containment introduces significant performance benefits and decreases rendering runtime. This is especially valuable to protect the performance of the website that integrates the Instant Expert. As of June 2020, CSS Containment is supported by default by the latest versions of major browsers (i.e. Google Chrome, Opera, Mozilla Firefox Nightly, and Microsoft Edge). Another advantage of using Content Containment is that it will not affect the functionality of Instant Expert even if the client browser does not support it.

**User Input**

There are three ways for a user to interact with the system; a) manually typing the question to a text box, b) invoking voice recognition to ask the question using a microphone, and c) selecting from a predefined list of questions (Figure 2). Text input is the most common type of interaction due to the search engine culture. Having a predefined list of questions allows the user to explore the system and better understand its capabilities. Voice-enabled communication is supported using Web Speech API, which is an experimental technology that defines a JavaScript API to integrate speech recognition and speech synthesis functionality into web pages. As of June 2020, speech recognition is supported by the latest versions of Google Chrome, Opera, and Microsoft Edge browsers. Speech synthesis is supported in all major browsers (e.g. Google Chrome, Mozilla Firefox, Opera, Microsoft Edge, and Safari). The component checks the client browser at initialization to test if Web Speech API is supported and disables speech features if not supported. The component allows the incorporation of third-party speech recognition and synthesis APIs, however, it requires modification of the component's source code. Upon the construction of the natural language question in text format, all input types eventually follow through the same flow to be passed to the knowledge generation module.

## 2.2 Knowledge Generation Module

The knowledge generation module is tasked with the complete workflow of producing a direct response to the given natural language question. There are two main approaches for powering the Instant Expert with a knowledge base; knowledge engine mode and question and answer (Q&A) mode. The knowledge engine mode relies on the integration of an external server-side knowledge engine that is capable of accessing and analyzing vast amounts of data in response to a natural language question. The Q&A mode requires a list of question & answer pairs supplied manually or by providing the URL of a webpage containing such pairs (e.g. Frequently Asked Questions). The Q&A mode utilizes deep learning models to create a tensor representation for each question in the knowledge base in order to calculate semantic similarity with the user question.

### 2.2.1 Q&A Mode

A major motivation of Instant Expert is to augment existing web platforms with a plug-and-play importable web component with minimal effort. In most use cases, static and textual responses can suffice to help users find useful information that they were looking for. Such pieces of information are often presented in a web platform in the form of Frequently Asked Questions (FAQ). However, searching for information via FAQs is often discouraging, hard to navigate, time-consuming, and results in failure of communicating critical information in a timely manner [33]. As a solution, the Q&A Mode of the Knowledge Generation Module is equipped with functionalities to parse and process FAQ pages and to efficiently and effectively map any user question into one of the question and answer pairs in the generated knowledge base. Thus, the Instant Expert effortlessly enhances user experience by allowing the users to verbally communicate with the system and receive a direct response without the hassle of going through potentially hundreds of frequently asked questions. Figure 2 visualizes the workflow for the Q&A Mode.

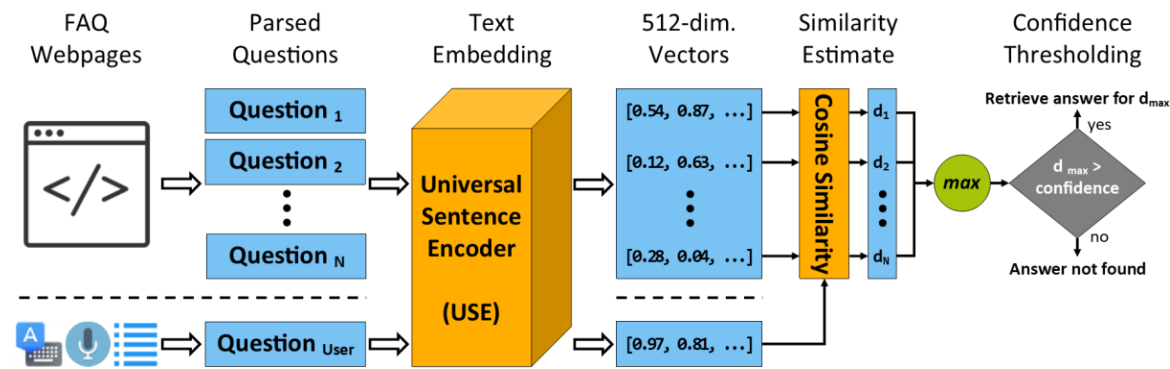

Figure 2: Flowchart summarizing the Q&A Mode's execution logic.

## Q&A Parsing and Collection

There are two approaches to collect Q&A pairs; manual entry and automatic FAQ parsing. For web platforms that do not have already established FAQ pages, the web component allows the developers to enter questions and their respective answers through HTML <slot> elements within the scope of Instant Expert in a hierarchical and organized manner. Since the Q&A pairs are directly provided, no further processing has been done before the encoding stage.

Automatic FAQ parsing requires the developer to supply the direct URL of the webpage containing the FAQ into the web component as an attribute. Both same-origin and cross-origin addresses are accepted given that the FAQ pages, that are not hosted in the same domain as the application, are navigated via a proxy server. The strength of the FAQ parsing and Q&A extraction comes from its unique design where no preprocessing is required regarding the webpage structure. Heuristic algorithms can perform just as well or better than machine learning approaches for similar scraping tasks as reported in the literature [34]. Our novel heuristic algorithm is built on the basis of two assumptions concerning the structure of the FAQ webpage: 1) the majority of question mark characters belong to the FAQ on a given page, and 2) both the questions and answers follow a certain HTML structure (i.e. not arbitrarily placed). The first assumption is clearly the case for the majority, if not all, of the FAQ pages since most FAQ questions end with a question mark and it is extremely unlikely to have more question marks in the remaining of the page. As for the second assumption, all questions in a given FAQ webpage is highly likely to follow a certain convention for the purpose of clarity, symmetry, maintainability, and effectiveness. More concretely, the HTML elements containing the FAQ question will likely share certain characteristics such as element tag, class, styling, immediate parents, and DOM depth. Thus, a pattern can be inferred by analyzing these correlations in an efficient way. The same condition is applicable for the FAQ answers.

The parsing process begins with finding innermost HTML elements containing a question mark to retrieve unique texts. Per our assumption, the majority of these question marks should belong to FAQ questions. To deduce a pattern, we traverse each retrieved element to keep record of their and their closest three parents, if exist, tag names along with their depth within the DOM tree. Per our symmetry assumption, the elements containing FAQ questions should produce the same values for tag names and depth. Using a hash table, the frequency of each combination is measured to identify the correct pattern. Thus, all FAQ questions are successfully retrieved based on the inferred pattern. Before initiating the process for parsing the FAQ answers, a challenge arises regarding the scope. For example, the innermost element containing the question might have parent elements for visual and interactive purposes (e.g. buttons, container boxes) before continuing to the following question or answer. It is known that the answer to a question lies somewhere between the text of that question and the text of next question. However, it cannot be known whether the answer is grouped with its question:

```html
<div id="questionGroup1">
    <div id="question1Container">...</div>
    <div id="answer1Container">...</div>
</div>
```

or simply have been coded sequentially in the same scope with other questions and answers:

```html
<div id="question1Container">...</div>
<div id="answer1Container">...</div>
```

Even if processing would have been performed to find the uppermost parent containing only the question text, scope issues still would not fully resolve since it is a common practice in FAQ pages to group questions by topic, which breaks the sibling relationship between the last question in a group and the first question in the next group. As a solution, a heuristic approach is taken to deduce a pattern to work in all possible webpage configurations. First, the uppermost parent of each question that does not contain its following question is retrieved, and its distance to the child is recorded. Per the symmetry assumption, the majority of retrievals will share the same distance value. Thus, the system is able to extract all answers regardless of their scope and structuring. Thus, all Q&A pairs are extracted from an FAQ webpage in a heuristic manner requiring no data other than the URL of the Q&A page.

**Q&A Encoding**

The extracted Q&A pairs are processed to create an encoded and numeric representation of the questions so that the user question can effectively be mapped. Due to the high variety of expression styles for questions sharing the same intent and desiring the same output, the structure and the words in the sentence should be analyzed while preserving the semantic integrity [35]. Furthermore, this representation will only be used to calculate similarity between different question sentences, and thus, natural language parsing and data extraction is not a necessity. An additional design choice is to be able to efficiently perform the encoding process on the client-side. Given these requirements, the Instant Expert utilizes the Universal Sentence Encoder (USE) [36], a model that is designed to embed text to produce a vector representation entailing the semantic information within the sentence. More specifically, the Transformer architecture [37] constitutes the underlying encoding model for context-aware representations of words resulting in a 512-dimensional vector. Within the presented framework, a lightweight version of the USE with an 8,000-word vocabulary is imported and executed with TensorFlow.js, an open-source JavaScript library that can train and deploy machine learning models on the client-side via the browser.

All questions (*n* questions) in the collected Q&A pairs are provided as input to the USE model via TensorFlow.js which produces a 512-dimensional tensor per question and results in a [*n*, 512]-sized matrix. This process takes place asynchronously in the background while allowing the users of the website to continue their interaction as usual. Once the processing is complete, the tensor matrix as well as the list of Q&A pairs are saved within the duration of the session. If the developer enabled *downloadModel* attribute of the Instant Expert web component, then the framework will generate a JSON file consisting of the tensor matrix and the Q&A pairs. This JSON file can be hosted on a server and the URL to access the file can be provided to Instant Expert. This mechanism allows the developers to prevent the client's hardware resources to unnecessarily be used by the embedding process, and make the Instant Expert instance to be initialized instantly to respond to any questions without needing to wait for the asynchronous operations to be completed. Thus, the framework allows three different usage styles for the Q&A Mode: same-origin or cross-origin FAQ webpage, manual Q&A definition, and JSON file containing Q&A pairs and question embeddings.

**Response Generation**

Each time a user asks a question through Instant Expert, the USE model is utilized in the same fashion to generate the embeddings in the form of a tensor. Thus, every predefined question with known answers as well as the input question are represented as points on a 512-dimensional coordinate system. Distance between these points is used as the criterion to assess semantic similarity. As the distance measurement technique, this framework utilizes the Cosine Similarity (Eq. 1) which is defined as;

$$\cos \theta = \frac{X.Y}{\|X\| \|Y\|} \tag{1}$$

, where X and Y are the tensor representations of the user's question and a predefined question in the parsed Q&A pairs, respectively. For robust calculation of dot products and norms, an open-source JavaScript library (Math.js) is utilized. This similarity is calculated for *n* questions that result in a similarity array where the values closer to 1 indicate higher correlation. Before completing the mapping to the question with the highest similarity, a threshold value is established to filter unrelated questions where a satisfactory answer is not present in the knowledge base. This is especially important to eliminate the potential to misguide users with unrelated or unappropriated information. This threshold value can be adjusted according to the nature of the use case to determine the trade-off between precision and recall. The framework performs the response generation phase asynchronously so that the hosting website is not throttled or disturbed. The generated response is returned to be displayed via the web component's interface along with the source of information to ensure the recipient of the information is aware of the source.

*2.2.2 Knowledge Engine Mode*

The presented component can be connected to an external knowledge engine for more sophisticated use cases of intelligent assistants. A server-side engine may utilize semantic webs to aid in the inference process and dynamic and distributed data resources to respond to complex queries. The main advantage of such engine is the use of custom natural language processing approaches capable of extracting useful information from the question such as time and date, location, intent, output type (e.g. graph, image, numeric value), ontological entities, and mathematical and statistical operations [38]. Thus, the knowledge generation process pertaining to an external engine has a computational essence in comparison to merely providing textual information.

The web component retrieves the answer for the input question by making an HTTP POST request to an external knowledge engine using the webhook link provided with the 'engine' attribute of the element. The only parameter passed to the engine is the question text using the parameter key 'question', which can be modified per developer's configuration. The component expects a response from the engine in JSON format with a key-value pair where value represents the response. The module requires the response to return within 2 seconds to portray a realistic and natural interaction with the user. If the engine and the web page that integrates the Instant Expert are not hosted on the same origin, then the engine should be able to handle requests from origins outside of its own by setting up Cross-Origin Resource Sharing (CORS), or utilize other solutions such as to use a proxy server or script.

## 3 RESULTS AND DISCUSSIONS

### 3.1 Q&A Mode – COVID-19 Case Study

The generalized nature of the presented framework makes it suitable for usage in any domain. For the purpose of demonstrating its workflow and benefits, an example use case needs to be selected where the necessity and impact of chatbots are evident. The degree of misinformation and sparseness of sources results in a data mess for topics ranging from politics to health. Social media and online news outlets often receive the same information from the same source, yet, reports it as duplication through a story with a narrative, which in turn, results in information pollution. One of the advantages of the Instant Expert is that it allows the sharing of information through numerous challenges by referencing the original source instead of duplication which opens the way for distrust. A recent example of such scenario is the ongoing pandemic of Coronavirus Disease 2019 (COVID-19). It has been widely reported how chatbots can prove to be useful during the pandemic on different levels envisioned and developed by various organizations and companies. Examples to these chatbot use cases regarding COVID-19 include, but not limited to, assessing eligibility for plasma donation [39], symptom checkers and health screening [40][41][42], and information dissemination, which is especially crucial as chatbots provide a direct and single response to a given question instead of the alternative of being succumbed to social media posts, and the spread of misleading or incomplete information [43][23]. It has been advised that local authorities and healthcare businesses should utilize chatbots to ensure 7/24 accurate information flow powered by the extensive amounts of Frequently Asked Questions available by authoritative sources such as CDC and WHO [44]. Yet, the realization of this vision is hindered by the lack of technological capabilities, resources, and funding available to such local and healthcare organizations. Thus, an information dissemination chatbot for COVID-19 has been selected as the use case in this paper to demonstrate the presented framework's usage and benefits due to the urgent demand as COVID-19 pandemic is progressing.

According to web analytics service [45], the CDC website has received the highest number of visits (i.e. traffic) among websites that are served in English and that offer information and statistics on the spread of the COVID-19 infection. Thus, we have selected the CDC's official Frequently Asked Questions webpage (https://www.cdc.gov/coronavirus/2019-ncov/faq.html) as a source for the following use cases. On that page, there is a total of 119 questions spanning various topics ranging from COVID-19 basics to cleaning and disinfection as of June 20, 2020. Figure 3 shows screenshots from the Instant Expert instance with CDC-powered Q&A set.

The case study has been performed on an average personal computer that is powered by Intel(R) Core(TM) i7-7700HQ CPU @ 2.80 GHz, 32 GB 2400 MHz RAM, and an NVIDIA GeForce GTX 1050.

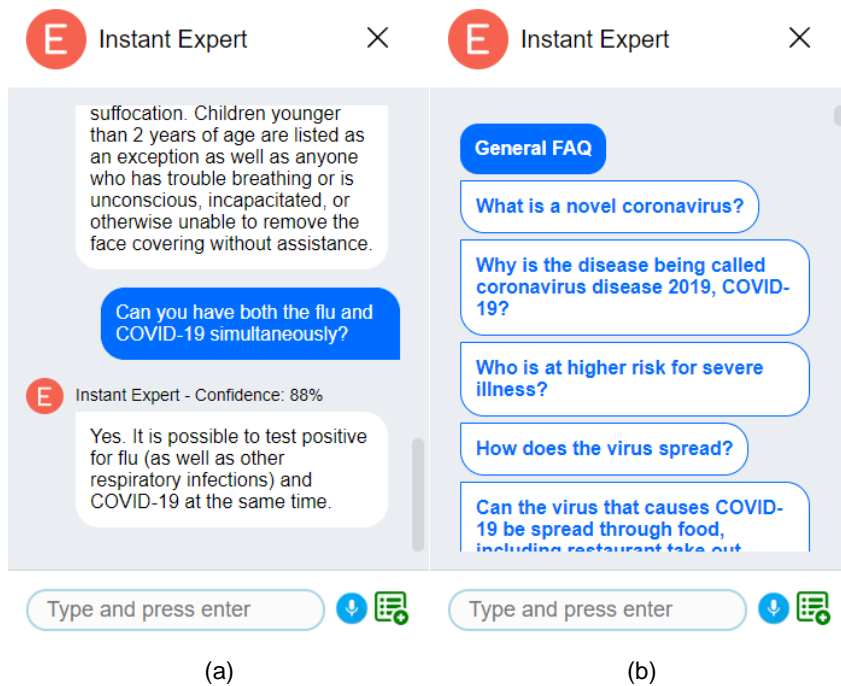

(a)                                            (b)

Figure 3: Screenshots from the Instant Expert for COVID-19 Chatbot. (a) The main screen of conversation with buttons to activate voice input and display example questions, (b) example questions that can be asked to the chatbot as recommendations to the user.

### 3.1.1 FAQ from a Web Page

The Instant Expert has been initialized in the FAQ mode and provided the CDC's FAQ URL as the information source. Due to the cross-origin limitations, a proxy server (CORS Anywhere) has been utilized to retrieve the webpage contents. Exactly all 119 Q&A items on the CDC page have successfully been extracted with a 100% precision and recall, followed by the embedding of all questions for natural language mapping. The average time spent on parsing the Q&A pairs and for embedding the questions using the Universal Sentence Encoder has been reported in Figure 4. In order to visualize the processing complexity, this benchmark has been performed for subsets of the CDC questions so that the correlation between the number of questions and the processing time is clear. The runtime measurement does not include the time spent for the retrieval of the FAQ page via GET request and loading the USE model as both of their performance is out of the presented framework's scope and affected by external parameters such as the internet speed. Furthermore, these actions are only performed once, and thus, is neither related to the number of questions nor the structure of the FAQ page.

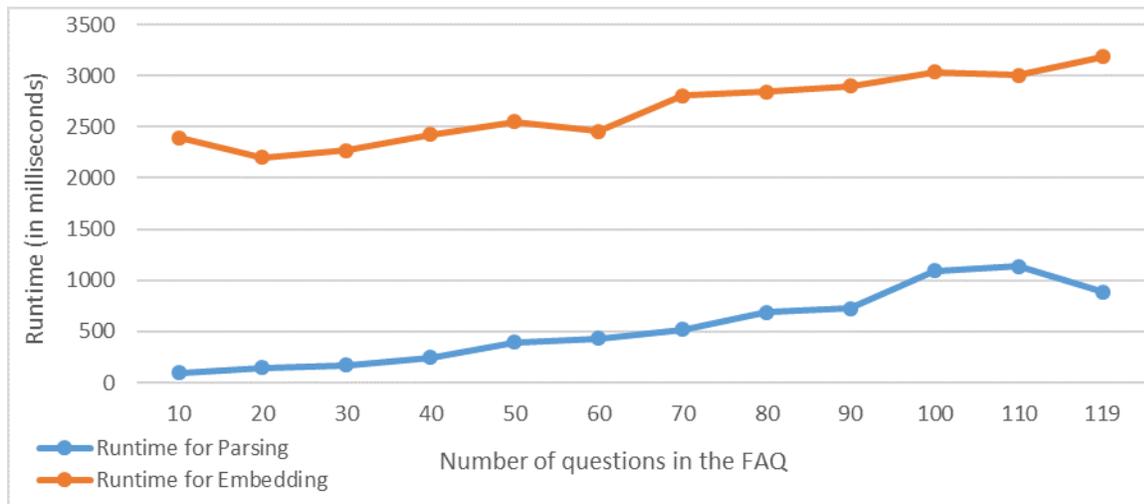

Figure 4: Benchmarks for powering the Instant Expert from an FAQ webpage: parsing time and embedding time.

One of the main purposes of the FAQ Mode is to allow users to ask questions in various natural language expressions. A test set is generated to quantify how flexible the Instant Expert is in terms of accurately mapping question variations that share the same intent. The test set contains the original FAQ question, answer, and one natural language question that expects the same answer with different phrasing. For objectivity, a third-party software (i.e. QuillBot), which is a machine learning-powered paraphraser and sentence restructurer, is utilized to produce high-variance natural language questions with a similar meaning to the original. Minor manual modifications have been made to questions in order to challenge the inference component of the system by referencing the actual questions asked by people on the internet. Table 1 provides examples of the generated questions in comparison to the original question. Additionally, the test set also contains three questions that the CDC's FAQ cannot and should not answer. These three questions are taken from the US Federal Drug Administration's (FDA) FAQ webpage. The reason for this addition is to make sure that the chatbot does not disseminate inaccurate and unrelated information. Thus, the test contains a total of 122 questions. For measurements, a benchmark analysis has been carried out to experiment with a broad range of confidence threshold values with respect to the precision and recall values. For reference, precision represents the percentage of questions that have been mapped to the correct answer out of all the questions that are mapped to some answer. Recall, on the other hand, represents the percentage of questions that have been accurately mapped. Both the benchmark code and a complete test set can be found on the GitHub repository for reproducibility and reanalysis.

Table 1: Example Questions from the Test Set

| Original Question | Generated Test Question |
|---|---|
| *"Should I make my own hand sanitizer if I can't find it in the stores?* | There are no hand sanitizer left in stores. Should I make one myself? |

| Original Question | Generated Test Question |
|---|---|
| *"What should I do if there is an outbreak in my community?* | What are you suggesting me to do if my community suffers from an outbreak? |
| *"Should I go to work if there is an outbreak in my community?* | Am I supposed to continue working if we have an outbreak in my street? |
| *"Can CDC tell me or my employer when it is safe for me to go back to work/school after recovering from or being exposed to COVID-19?"* | If I am exposed to COVID-19, when can I safely go back to work? |
| *"My family member died from COVID-19 while overseas. What are the requirements for returning the body to the United States?"* | What is the policy on bringing my relative back to US who passed away due to COVID-19? |
| *"What is routine cleaning? How frequently should facilities be cleaned to reduce the potential spread of COVID-19?"* | How often should I clean my place to prevent COVID-19? |
| *"What should I do if there are pets at my long-term care facility or assisted living facility?"* | What steps I should take if my nursing home has pets? |
| *N/A (CDC's FAQ does not have this question)* | Am I at risk for serious complications from COVID-19 if I smoke cigarettes? |
| *N/A (CDC's FAQ does not have this question)* | Are there any vaccines to prevent COVID-19? |
| *N/A (CDC's FAQ does not have this question)* | Are antibiotics effective in preventing or treating COVID-19? |

In order to quantify the model's effectiveness, precision (Eq. 2), recall (Eq. 3), and f1-score (Eq. 4) metrics have been selected for this imbalanced classification problem with multiple classes as formulated below [46]. $n$ value in the equations below represents the number of different questions in the FAQ (i.e. classes).

$$precision\ (multiclass) = \frac{\sum_0^n TruePositive}{\sum_0^n TruePositive + \sum_0^n FalsePositive} \quad (2)$$

$$recall\ (multiclass) = \frac{\sum_0^n TruePositive}{\sum_0^n TruePositive + \sum_0^n FalseNegative} \quad (3)$$

$$f1-score\ (multiclass) = \frac{2 * precision * recall}{precision + recall} \quad (4)$$

Figure 5 shows how precision and recall values are affected based on the selected confidence threshold. To achieve an accurate and complete system, the goal is to maximize the precision and recall values, however, a trade-off evaluation is required [47]. Based on the use case, it may be important to maximize the precision for making sure to always provide the accurate answer for questions that are mapped or to maximize recall for mapping as much question as possible while potentially sacrificing the accuracy of a few. For this use case, the recall starts to drastically decrease around confidence level of 0.8 while the precision stabilizes as seen on Figure 5. Looking at a confidence level of 0.75, the precision is maximized with a value of 100% while recall value is 97.541%. Coincidentally, the ideal confidence value for the highest possible recall is also observed to be 0.75, though for other cases this value might differ resulting in a trade-off to prioritize either maximum precision or recall. This ideal confidence level can also be observed with the harmonic mean of precision and recall values (i.e. f1-score) where f1-score is the highest at 0.75 confidence.

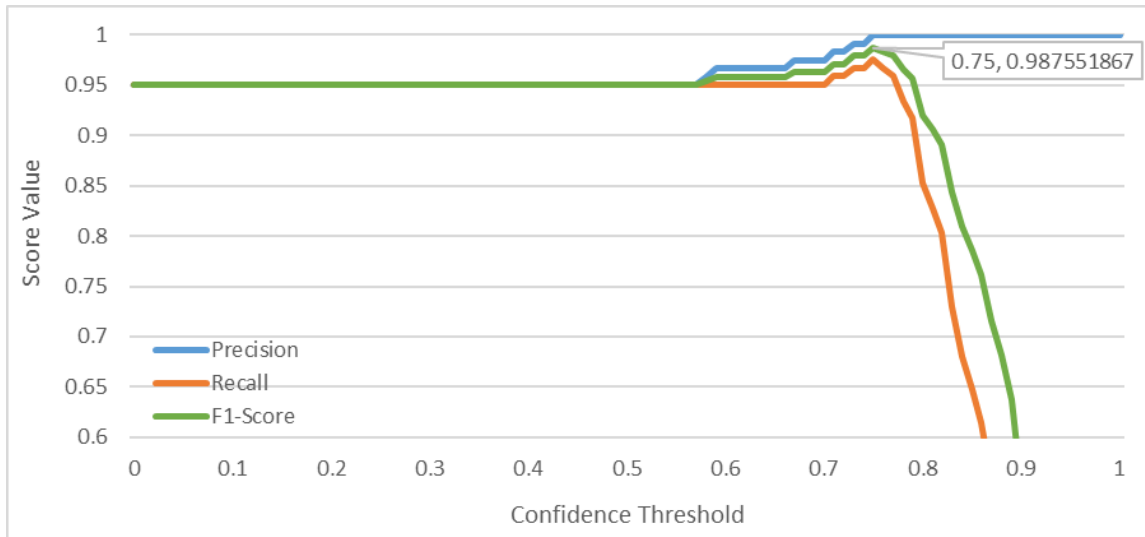

Figure 5: Precision, recall, and f1-score values for different confidence thresholds based on generated test data.

For applications when the prediction results in critical information regarding human health, such as the use case for COVID-19, precision is more critical than recall. Because, providing the wrong information can have detrimental results. Thus, based on the experiment, the confidence threshold is set to 0.75 by default to ensure perfect precision of 100%, which results in 3 out of 119 questions are left unmapped. These three questions are examined to identify why they have not reached to satisfactory confidences. Table 2 summarizes the original FAQ text, the test question, and commentary on why the mapping was not successful. Based on the investigation, we conclude the questions that are not context-independent and not well-structured are more difficult to be mapped from to-the-point questions that an average user of the system might ask.

Table 2: Summary of unmapped questions.

| Original Question | Test Question | Commentary |
|---|---|---|
| *"Limit time with older adults, including relatives, and people with chronic medical conditions."* | *"Should I avoid spending time with the elderly, especially those with health conditions?"* | The structure of the original question is not in the form of a question, and rather a recommendation. Especially since there are similar questions exist in the FAQ on dealing with people with underlying conditions, the mapping process couldn't complete with a satisfactory confidence. |

| Original Question | Test Question | Commentary |
|---|---|---|
| *"Will businesses and schools close or stay closed in my community and for how long? Will there be a "stay at home" or "shelter in place" order in my community?"* | *"For how much longer the business will stay closed?"* | This FAQ item is too broad, and in fact entails three different questions for business or schools. The test question only asks a portion of the FAQ item concerning the businesses, which results in unsatisfactory confidence. |
| *"What about imported animals or animal products?"* | *"Do animal products pose risk?"* | This FAQ question is incomplete as its meaning depend on a previous question in the FAQ list. |

The Instant Expert can be instantiated based on a FAQ webpage by simply providing the URL. This use case is especially suitable for cases where the information in the FAQ change frequently. As an example, it has been widely reported that the constantly and rapidly changing nature of COVID-19-related data and guidelines cause misinformation or confusion of the public [48]. Thus, this mode can justify the required client-side workload when the source is an ever-expanding and ever-changing list of question and answers. Example usage is shown below.

```
<instant-expert
        mode="faq-web"
        faq-url="YourProxyURL/
        https://www.cdc.gov/coronavirus/2019-
        ncov/faq.html">
</instant-expert>
```

**Parsing FAQs from Other Sources**

As a tangent to the COVID-19 Case Study that is powered by CDC data, we have experimented with FAQ webpages from other websites and domains to showcase the Instant Expert's generalized structure and domain independence. Table 3 summarizes five FAQ sources that were used in the same manner as described in the implementation above by providing the respective URL. The output of the parsing component has been recorded and analyzed to measure the success rate. In this context, precision represents the number of the accurate text groups that were labeled as a question or an answer, whereas, the recall represents out of all the question and answers on the given webpage the framework was able to accurately parse. The analyzed FAQ webpages are structured in various ways and in different languages (e.g. English, German, Spanish, French), thus, demonstrating the parsing process' generalized workflow. All FAQ pages below analyzed as of June 26th, 2020.

Table 3: Analysis of FAQs to demonstrate the framework's domain independence and generalized nature

| FAQ Source | | No of Q&A | No of Parsed Q&A | Precision | Recall |
|---|---|---|---|---|---|
| FDA COVID-19 FAQ (fda.gov) | | 78 | 78 | | |
| World Health Organization (WHO) – Q&A on coronaviruses (who.int) | | 24 | 24 | 100& | 100% |
| United Nations COVID-19 FAQ (un.org) | (in English) | 40 | 40 | | |
| | (in French) | 37 | 37 | | |

| FAQ Source | | No of Q&A | No of Parsed Q&A | Precision | Recall |
|---|---|---|---|---|---|
| | (in Spanish) | 38 | 38 | | |
| Stanford COVID-19 FAQ (stanfordhealthcare.org) | | 16 | 16 | | |
| Robert Koch Institut COVID-19 FAQ (rki.de) | (in German) | 43 | 43 | | |

### 3.1.2  FAQ from a Custom List

Some use cases may require manual definition of question and answers instead of having or relying on an existing FAQ webpage. To enable such initialization, the Instant Expert presents a mode, called *faq-custom*, in which HTML Slot elements are utilized to allow the developer to specify questions and their corresponding answers as shown below.

```
<instant-expert mode="faq-custom">
    <div slot="questions">
        <p>Question 1</p>
        <p>Question 2</p>
    </div>
    <div slot="answers">
        <p>Answer 1</p>
        <p>Answer 2</p>
    </div>
</instant-expert>
```

### 3.1.3  FAQ from a Model

Most of the parsing and embedding process takes place on the background (i.e. async) to allow users to continue normal operation, however, it still consumes client resources and requires varying time depending on client hardware. Both previous FAQ processing approaches (i.e. web, custom) come with the capability of extracting a JSON file containing the parsed Q&A pairs along with their USE embeddings (e.g. a 512-dimensional tensor). This downloaded *model* file can be provided to the Instant Expert directly to eliminate the time and resources required for FAQ processing. This use case is suggested as the default method to ensure the users can use the chatbot immediately after the page is loaded. Since no new processing is done, the precision and recall values are the same as reported above. Example usage is presented below.

```
<instant-expert
    mode="faq-model"
    faq-url="instantexpert_faq_cdc.json">
</instant-expert>
```

## 3.2  Knowledge Engine Mode

An example use case of the presented framework in *Knowledge Engine Mode* is the development of a generic question answering assistant using the *Project Answer Search* by Microsoft Cognitive Labs. Project Answer Search is an experimental technology to instantly answer natural language user queries with factual responses [49]. This use case is specifically developed to be a complete solution that (a) will serve as an adoption guide to the users, (b) can be easily accessed to try the presented software, (c) and can be conveniently reproduced for production use or ensure the component's correctness. There are two parts constituting this use case; developing a question-answering engine as the backend and developing a website that implements the Instant Expert component integrated with the engine. The backend is implemented as a Node.js application and served on a cloud platform (i.e. Heroku). The source codes for both the backend and the website are available on Instant Expert GitHub repository in the examples directory along with the directions necessary for reproduction. The validation of use cases of the Knowledge Engine Mode belongs to the external natural-language question answering service where the accuracy can be measured for response generation. For this use case, an already validated system (i.e. Project Answer Search) is utilized to focus on the framework's communication mechanism. Figure 6 shows this use case in action.

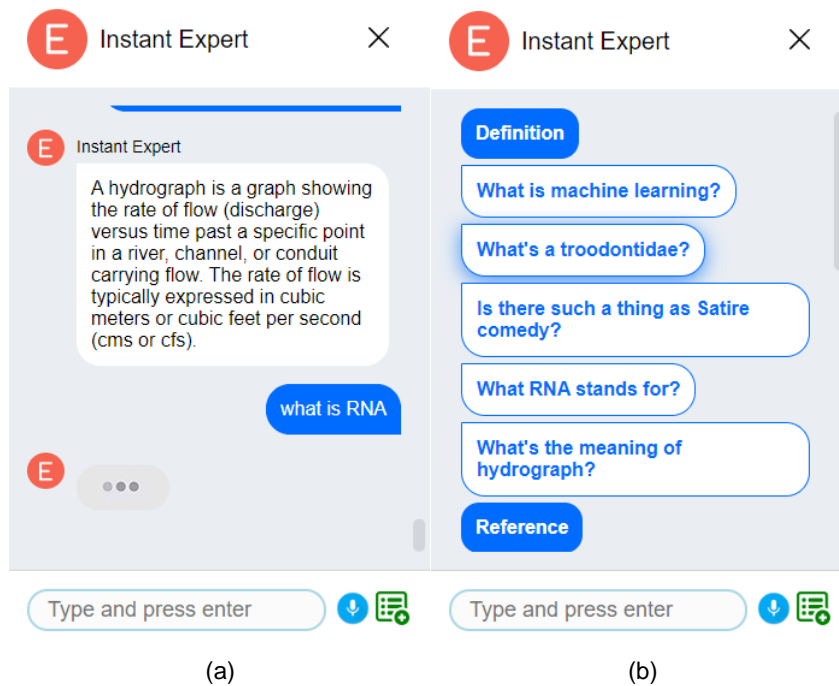

Figure 6: Screenshots from the Instant Expert powered by an external knowledge engine.

## 4 CONCLUSIONS

This research introduces the Instant Expert, an open-source semantic web framework to effortlessly create and integrate fully functional voice-enabled smart assistants (i.e. chatbots) into any web platform with as little as a single line of HTML code. It provides a complete solution with its user interface and functionalities, communication protocols, and knowledge generation components powered by heuristic algorithms and deep learning models. It utilizes a variety of advanced web technologies including Web Components with Shadow DOM and CSS Containment to provide an isolated, robust, and efficient solution, and frees the developers from the complications of integrating third-party components into existing web platforms.

To the best of our knowledge, the presented framework is the first FAQ-powered chatbot that is fully open-source and free-to-use as well as the first that can fully operate on the client-side, which is favorable to minimize maintenance and operation costs and to protect user data. Additionally, the framework minimizes the complexity and the effort needed to implement and maintain chatbots and allow individual developers, academic research groups, small companies, non-profits, and public offices to enrich their web platforms with natural language communication. Furthermore, information delivery enhanced with natural language voice interactions instead of browsing through broad and entangled websites simplify the learning process for people with cognitive and mobility impairments. Another potential advantage is that the simplified design of the Instant Expert's visual elements eliminates distractions in the learning process for people with attention-deficit/hyperactivity disorder (ADHD).

For future work, there are many paths to advance the proposed vision. The parsing component of the FAQ Mode can be improved to 1) recursively parse question pages with links and 2) to group questions by their focus to recommend solutions to the user in cases where their question cannot directly be answered. Another potential improvement to the FAQ Mode would be to allow multiple custom question variations to better model an intent so that the recall can be maximized while the knowledge base scales. For the Knowledge Engine Mode, integration modules can be developed to effortlessly connect third-party services for natural language processing and inference. The Instant Expert's scope and capabilities can be expanded into virtual reality (e.g. A-Frame), and augmented reality (i.e. HoloLens and Magic Leap) applications. Finally, the framework can be utilized to scrape multiple FAQs from a variety of sources to create a smart assistant for a selected subject in any domain.

## ACKNOWLEDGMENTS

This project is based upon work supported by the Iowa Flood Center and the University of Iowa.

## REFERENCES

[1] Wu, X., Zhu, X., Wu, G.Q. and Ding, W., 2013. Data mining with big data. *IEEE transactions on knowledge and data engineering*, 26(1), pp.97-107.

[2] Sermet, Y., Villanueva, P., Sit, M.A. and Demir, I., 2020. Crowdsourced approaches for stage measurements at ungauged locations using smartphones. *Hydrological Sciences Journal*, 65(5), pp.813-822.

[3] Sit, M., Sermet, Y. and Demir, I., 2019. Optimized watershed delineation library for server-side and client-side web applications. *Open Geospatial Data, Software and Standards*, 4(1), p.8.

[4] Demir, I., Yildirim, E., Sermet, Y. and Sit, M.A., 2018. FLOODSS: Iowa flood information system as a generalized flood cyberinfrastructure. *International journal of river basin management*, 16(3), pp.393-400.

[5] Peppard, J. and Ward, J., 2016. *The strategic management of information systems: Building a digital strategy*. John Wiley & Sons.

[6] Carson, A., Windsor, M., Hill, H., Haigh, T., Wall, N., Smith, J., Olsen, R., Bathke, D., Demir, I. and Muste, M., 2018. Serious gaming for participatory planning of multi-hazard mitigation. International journal of river basin management, 16(3), pp.379-391.

[7] Alberts, I., 2013. Challenges of information system use by knowledge workers: The email productivity paradox. *Proceedings of the American Society for Information Science and Technology*, 50(1), pp.1-10.

[8] Sermet, Y. and Demir, I., 2018. An intelligent system on knowledge generation and communication about flooding. *Environmental modelling & software*, 108, pp.51-60.

[9] Xu, H., Windsor, M., Muste, M. and Demir, I., 2020. A web-based decision support system for collaborative mitigation of multiple water-related hazards using serious gaming. *Journal of Environmental Management*, 255, p.109887

[10] Sermet, Y., Demir, I. and Muste, M., 2020. A serious gaming framework for decision support on hydrological hazards. *Science of The Total Environment*, p.138895

[11] Liao, S.H., 2005. Expert system methodologies and applications—a decade review from 1995 to 2004. *Expert systems with applications*, 28(1), pp.93-103.

[12] Brandtzaeg, P.B. and Følstad, A., 2017, November. Why people use chatbots. In *International Conference on Internet Science* (pp. 377-

392). Springer, Cham.

[13] Sit, M.A., Demiray, B.Z., Xiang, Z., Ewing, G., Sermet, Y. and Demir, I., 2020. A Comprehensive Review of Deep Learning Applications in Hydrology and Water Resources.

[14] IMARC Group, 2019, June. *Intelligent Virtual Assistant Market: Global Industry Trends, Share, Size, Growth, Opportunity and Forecast 2019-2024* (Report ID: 4775648). https://www.researchandmarkets.com/reports/4775648/intelligent-virtual-assistant-market-global

[15] Schoemaker, P.J. and Tetlock, P.E., 2017. *Building a more intelligent enterprise.* MIT Sloan Management Review.

[16] Jain, M., Kumar, P., Kota, R. and Patel, S.N., 2018, June. Evaluating and informing the design of chatbots. In *Proceedings of the 2018 Designing Interactive Systems Conference* (pp. 895-906).

[17] Chung, K. and Park, R.C., 2019. Chatbot-based heathcare service with a knowledge base for cloud computing. *Cluster Computing, 22*(1), pp.1925-1937.

[18] Sermet, Y. and Demir, I., 2019. Flood action VR: a virtual reality framework for disaster awareness and emergency response training. In ACM SIGGRAPH 2019 Posters (pp. 1-2).

[19] Yildirim, E. and Demir, I., 2019. An integrated web framework for HAZUS-MH flood loss estimation analysis. Natural Hazards, 99(1), pp.275-286.

[20] Androutsopoulou, A., Karacapilidis, N., Loukis, E. and Charalabidis, Y., 2019. Transforming the communication between citizens and government through AI-guided chatbots. *Government Information Quarterly, 36*(2), pp.358-367.

[21] Vaidyam, A.N., Wisniewski, H., Halamka, J.D., Kashavan, M.S. and Torous, J.B., 2019. Chatbots and conversational agents in mental health: a review of the psychiatric landscape. *The Canadian Journal of Psychiatry, 64*(7), pp.456-464.

[22] USACE, 2019. Virtual Assistant Technology Holds Promise for USACE. *Engineer Update – The Official Newsletter of the U.S. Army Corps of Engineers*, 8 November. Alexandria, Virginia.

[23] Miner, A.S., Laranjo, L. and Kocaballi, A.B., 2020. Chatbots in the fight against the COVID-19 pandemic. *npj Digital Medicine, 3*(1), pp.1-4.

[24] Sohrabi, C., Alsafi, Z., O'Neill, N., Khan, M., Kerwan, A., Al-Jabir, A., Iosifidis, C. and Agha, R., 2020. World Health Organization declares global emergency: A review of the 2019 novel coronavirus (COVID-19). *International Journal of Surgery.*

[25] Schmidt, B., Borrison, R., Cohen, A., Dix, M., Gärtler, M., Hollender, M., Klöpper, B., Maczey, S. and Siddharthan, S., 2018, October. Industrial Virtual Assistants: Challenges and Opportunities. In *Proceedings of the 2018 ACM International Joint Conference and 2018 International Symposium on Pervasive and Ubiquitous Computing and Wearable Computers* (pp. 794-801).

[26] Daniel, G., Cabot, J., Deruelle, L. and Derras, M., 2020. Xatkit: a multimodal low-code chatbot development framework. *IEEE Access, 8,* pp.15332-15346.

[27] Mind Commerce, 2019. *Virtual Personal Assistants (VPA) and Smart Speaker Market: Artificial Intelligence Enabled Smart Advisers, Intelligent Agents, and VPA Devices 2019 – 2024.* https://mindcommerce.com/reports/virtual-personal-assistant-market/

[28] Adam, M., Wessel, M. and Benlian, A., 2020. AI-based chatbots in customer service and their effects on user compliance. *Electronic Markets,* pp.1-19.

[29] Sermet, Y. and Demir, I., 2020. Virtual and augmented reality applications for environmental science education and training. *New Perspectives on Virtual and Augmented Reality: Finding New Ways to Teach in a Transformed Learning Environment.*

[30] Oh, J., Ahn, W.H. and Kim, T., 2017, November. Web page restructuring based on shadow DOMs to improve maintainability. In *2017 8th IEEE International Conference on Software Engineering and Service Science (ICSESS)* (pp. 118-122). IEEE.

[31] De Ryck, P., Nikiforakis, N., Desmet, L., Piessens, F. and Joosen, W., 2015. Protected web components: Hiding sensitive information in the shadows. *IT Professional, 17*(1), pp.36-43.

[32] Atkins, T., Rivoal, F. CSS Containment Module Level 1. 18 November 2018. WD. URL: https://www.w3.org/TR/css-contain-1/

[33] Damani, S., Narahari, K.N., Chatterjee, A., Gupta, M. and Agrawal, P., 2020, May. Optimized Transformer Models for FAQ Answering. In *Pacific-Asia Conference on Knowledge Discovery and Data Mining* (pp. 235-248). Springer, Cham.

[34] Jijkoun, V. and de Rijke, M., 2005, October. Retrieving answers from frequently asked questions pages on the web. In *Proceedings of the 14th ACM international conference on Information and knowledge management* (pp. 76-83).

[35] Farouk, M., 2020. Measuring Text Similarity Based on Structure and Word Embedding. *Cognitive Systems Research.*

[36] Cer, D., Yang, Y., Kong, S.Y., Hua, N., Limtiaco, N., John, R.S., Constant, N., Guajardo-Cespedes, M., Yuan, S., Tar, C. and Sung, Y.H., 2018. Universal sentence encoder. *arXiv preprint arXiv:1803.11175.*

[37] Vaswani, A., Shazeer, N., Parmar, N., Uszkoreit, J., Jones, L., Gomez, A.N., Kaiser, Ł. and Polosukhin, I., 2017. Attention is all you need. In *Advances in neural information processing systems* (pp. 5998-6008).

[38] Sermet, Y. and Demir, I., 2019. Towards an information centric flood ontology for information management and communication. *Earth Science Informatics, 12*(4), pp.541-551.

[39] CNBC, 2020, April. *Microsoft is launching a 'plasmabot' to encourage people who recovered from the virus to donate their plasma as a possible treatment.* https://www.cnbc.com/2020/04/18/microsoft-plasmabot-encourages-covid-19-survivors-to-donate-plasma.html [accessed 15 June 2020].

[40] Espinoza, J., Crown, K. and Kulkarni, O., 2020. A Guide to Chatbots for COVID-19 Screening at Pediatric Health Care Facilities. *JMIR Public Health and Surveillance, 6*(2), p.e18808.

[41] Judson, T.J., Odisho, A.Y., Young, J.J., Bigazzi, O., Steuer, D., Gonzales, R. and Neinstein, A.B., 2020. Case Report: Implementation of a Digital Chatbot to Screen Health System Employees during the COVID-19 Pandemic. *Journal of the American Medical Informatics Association.*

[42] Martin, A., Nateqi, J., Gruarin, S., Munsch, N., Abdarahmane, I. and Knapp, B., 2020. An artificial intelligence-based first-line defence against COVID-19: digitally screening citizens for risks via a chatbot. *bioRxiv.*

[43] Sharma, M., Yadav, K., Yadav, N. and Ferdinand, K.C., 2017. Zika virus pandemic—analysis of Facebook as a social media health information platform. *American journal of infection control*, *45*(3), pp.301-302.

[44] Vergadia, P., 2020. How can Chatbots help during global pandemic (COVID-19)? *Google Cloud Community Articles.*

[45] SimilarWeb, "Coronavirus Data, Insights, and Trends," SimilarWeb, New York, NY, USA, 2020, https://www.similarweb.com/coronavirus/

[46] Sokolova, M. and Lapalme, G., 2009. A systematic analysis of performance measures for classification tasks. *Information processing & management*, *45*(4), pp.427-437.

[47] He, H. and Ma, Y. eds., 2013. *Imbalanced learning: foundations, algorithms, and applications.* John Wiley & Sons.

[48] Ross, C., 2020. I asked eight chatbots whether I had Covid-19. The answers ranged from 'low' risk to 'start home isolation' *STAT Health Tech Newsletter.*

[49] Microsoft Cognitive Services Lab (2018). Project Answer Search, https://labs.cognitive.microsoft.com/en-us/project-answer-search, [accessed 15 June 2020].